\documentclass[11pt,a4paper]{article}
\usepackage[hyperref]{emnlp2020}
\usepackage{times}
\usepackage{latexsym}

\usepackage{microtype}

\usepackage{listings}
\usepackage{amsmath}
\usepackage{colortbl}
\usepackage{array,multirow,graphicx}

\usepackage{graphicx}
\usepackage{subcaption}
 
\usepackage{url}
\aclfinalcopy 

\newcommand{\scell}[1]{\multicolumn{1}{r|}{\scriptsize{#1}}}
\newcommand{\srcell}[1]{\multicolumn{1}{r}{\scriptsize{#1}}}
\newcommand{\sgcell}[1]{\multicolumn{1}{r|}{\cellcolor{Gray}{\scriptsize{#1}}}}
\newcommand{\sgrcell}[1]{\multicolumn{1}{r}{\cellcolor{Gray}{\scriptsize{#1}}}}

\definecolor{Gray}{gray}{0.9}
\newcolumntype{a}{>{\columncolor{Gray}}c}

\title{Complete Multilingual Neural Machine Translation}

\author{Markus Freitag and Orhan Firat\\
  Google Research \\
  {\tt \{freitag,orhanf\}@google.com}}

\date{}

\begin{document}
\maketitle
\begin{abstract}
Multilingual Neural Machine Translation (MNMT) models are commonly trained on a joint set of bilingual corpora which is acutely English-centric (i.e. English either as the source or target language). While direct data between two languages that are non-English is explicitly available at times, its use is not common. In this paper, we first take a step back and look at the commonly used bilingual corpora (WMT), and resurface the existence and importance of implicit structure that existed in it: multi-way alignment across examples (the same sentence in more than two languages). We set out to study the use of multi-way aligned examples to enrich the original English-centric parallel corpora. We reintroduce this direct parallel data from multi-way aligned corpora between all source and target languages. By doing so, the English-centric graph expands into a complete graph, every language pair being connected. We call MNMT with such connectivity pattern complete Multilingual Neural Machine Translation (cMNMT) and demonstrate its utility and efficacy with a series of experiments and analysis. In combination with a novel training data sampling strategy that is conditioned on the target language only, cMNMT yields competitive translation quality for all language pairs. We further study the size effect of multi-way aligned data, its transfer learning capabilities and how it eases adding a new language in MNMT. Finally, we stress test cMNMT at scale and demonstrate that we can train a cMNMT model with up to 111$*$112=12,432 language pairs that provides competitive translation quality for all language pairs.

\end{abstract}

\section{Introduction}
Multilingual machine translation \cite{dong-etal-2015-multi,firat2016multi,johnson2017google,aharoni2019massively}, which can serve multiple language pairs with a single model, has attracted much attention. In contrast to bilingual MT systems which can only serve one single language pair, multilingual models can serve $O(N^2)$ language pairs ($N$ being the number of languages in a multilingual model) \cite{zhang2020improving}.

\begin{figure}
    \centering
    \begin{subfigure}[b]{0.17\textwidth}
        \includegraphics[width=\textwidth]{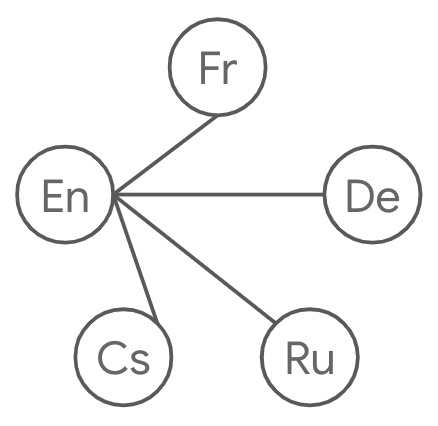}
        \caption{English-centric}
        \label{fig:en-centric}
    \end{subfigure}
    \quad 
    \begin{subfigure}[b]{0.18\textwidth}
        \includegraphics[width=\textwidth]{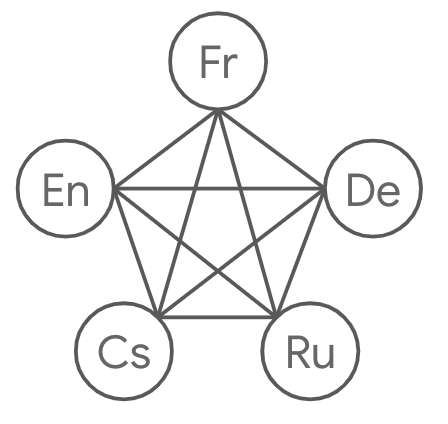}
        \caption{Complete}
        \label{fig:complete}
    \end{subfigure}    
    \caption{Source-target translation graphs in MNMT. Solid lines indicate that there exist direct parallel data. When there is no line connecting any two languages, zero-resource or zero-shot approaches are employed.}
    \label{fig:graph}
    \vspace{-1.5em}
\end{figure}

The amount of available training data can differ a lot across language pairs and the majority of available MT training data is English-centric \cite{DBLP:journals/corr/abs-1802-00273,DBLP:journals/corr/abs-1907-05019} which in practice means that most non-English language pairs do not see a single training example when training multilingual models (see Figure~\ref{fig:en-centric}). 
As a consequence, the actual performance of language pairs that do not include English on the source or target side lags behind the ones with large amounts of training data. Further, when increasing the number of languages, it gets (a) impractical to gather training data for each language pair and (b) challenging to find the right mix during training. Which is why models tasked with direct translation between non-English pairs either resort to bridging (pivoting) through a pivot language \cite{habash-hu-2009-improving}, or make use of synthetic parallel data (via back-translation) \cite{firat2016zero,chen2017teacher} or study the problem under zero-shot settings \cite{johnson2017google,ha2016toward}. 

In this study, we make use of the potential pre-existing multi-way property in the training corpora and generate as many direct training examples from pre-existing English-centric training data. If we can find training examples for each language pair in a multilingual mix, we call this model complete Multilingual Neural Machine Translation (cMNMT). cMNMT is then trained on all bilingual pairs between source and target languages by utilizing multi-way aligned training examples that consist of translations of the same sentence into multiple languages. We resurface multi-way aligned training examples by aligning training examples from different language pairs when either their source or target sides are identical (ie. pivoting through English, for German$\to$English and English$\to$French to extract German--French--English examples).

To make use of this data, the model samples a source and target language from the set of multi-way aligned corpus during training, which allows the model to see language pairs where originally no training data existed (missing connections in Figure~\ref{fig:en-centric}). As our experiments support, this method enables us to get access to training data for all tested language pairs (generating a complete graph (Figure~\ref{fig:complete})). We will show that it is possible to generate a complete graph for at least a 6-language WMT setup. Some of the WMT training data is multi-way parallel by construction. Nevertheless, we show that we also find many training examples where the source and target origin from different sources. We further show on our 112 languages internal dataset, that we can find sufficient training data for over 12,000 language pairs by only providing 111 English-centric training corpora. This result indicates that it is possible to generate direct training data for many language pairs without the need for crawling new training examples. Our experiments suggest that before falling back to methods like zero-shot translation, you should investigate the structure of your pre-existing training data.

To address the problem of finding the right mix of examples from different language pairs during training, we further introduce a hierarchical sampling strategy that is language-specific (as opposed to being language pair specific). In addition to fixing some chronic issues of MNMT (i.e. low quality for out of English translation \cite{firat2016multi,johnson2017google,DBLP:journals/corr/abs-1907-05019}), the proposed sampling strategy efficiently ensures all source-target pairs are covered.

Experiments demonstrate that we can train a cMNMT model on a 30-language-pair WMT setup that outperforms bilingual and multilingual baselines as well as bridging on all non-English language pairs. We further show that the performance of the English language pairs stay stable and do not suffer from the changes in both the training data and the new training data sampling strategy.
Furthermore, we share experiments at scale by demonstrating that we can train a cMNMT model that can serve  12,432 language pairs.

Our contribution is three-fold:

\begin{itemize}
    \item We show that we can find a lot of training examples for all language pairs in a multilingual mix by only pivoting pre-existing English-centric training data. We further show that many of the extracted examples originate from different data sources and this method could scale to many more datasets. We also support these findings with experiments on our internal dataset, where we were able to find training data for all 12,432 language pairs.
    \item We demonstrate that cMNMT outperforms bilingual baselines, multilingual baselines as well as bridging on all non-English language pairs while keeping translation performance on English-centric language pairs.
    \item We introduce a new sampling strategy that is purely based on the target language instead of language pairs and does scale to MNMT models which hundreds of languages.
\end{itemize}

\section{A Peek at Multi-way Aligned Examples in Bilingual Corpora}
\label{subsec:wmt_nway}
We choose six languages Czech (cs), English (en), French (fr), German (de), Spanish (es) and Russian (ru) from the public WMT datasets. The selection of the languages was driven by the fact that the WMT 2013 evaluation campaign~\cite{bojar-etal-2013-findings} released a multi-way test set for these six languages. 
As training data, we used WMT 2013 for Spanish, WMT 2014 for German, WMT 2015 for French, and WMT 2018 for Czech and Russian. 

We can construct non-English bilingual training examples by pairing the non-English sides of two training examples with identical English translations.
Table~\ref{table:wmt_training_data_stats} shows the number of bilingual training examples that we could potentially extract from the English-centric training data. The number of training examples for each non-English language pair varies from at least 0.3 million (Russian-German) to up to 4.8 million sentence pairs (Russian-French).

\begin{table}[ht]
\begin{center}
{\setlength{\tabcolsep}{.23em}
\begin{tabular}{l||c|c|a|c|c|c}
 & cs & de & en & es & fr & ru\\ \hline \hline
cs &  & 0.7 & 47 & 0.8 & 1 & 0.9 \\ \hline
de & 0.7 &  & 4.5 & 2.3 & 2.5 & 0.3 \\ \hline
\rowcolor{Gray}
en & 47 & 4.5 &  & 13.1 & 38.1 & 33.5 \\ \hline
es & 0.8 & 2.3 & 13.1 &  & 10 & 4.4 \\ \hline
fr & 1 & 2.5 & 38.1 & 10 &  & 4.8 \\ \hline
ru & 0.9 & 0.3 & 33.5 & 4.4 & 4.8 &  \\
\end{tabular}
}
\end{center}
\vspace{-0.5em}
\caption{WMT: Available training data (in million) after constructing non-English examples from English-centric examples with identical English side.}
\label{table:wmt_training_data_stats}
\end{table}

\begin{table*}[ht]
\begin{center}
{\setlength{\tabcolsep}{.23em}
\begin{tabular}{l||c|c|c|c|c|c|c|c|c}
& cc & CzEng & epps & nc & $10^9$ & Paracrawl & UN & Wiki Titles & Yandex \\ \hline \hline
Common Crawl (cc) & \cellcolor{Gray}{2.5M} & 13K & 213 & 21k & 10k & 47k & 1.8k & 1 & 6.1k \\ \hline 
CzEng 1.7 & & 0 & 20k & 417k & 242k & 55k & 63k & 98k & 7.7k \\ \hline
Europarl (epps) & & & \cellcolor{Gray}{6.9M} & 1.2k & 3.7k & 4.8k & 4.7k & 255 & 280 \\ \hline
News-Commentary (nc) & & & & \cellcolor{Gray}{640k} & 186k & 244 & 305 & 60 & 1.7k \\ \hline
$10^9$ & & & & & 0 & 12k & 97k & 1.5k & 2.9k \\ \hline
Paracrawl & & & & & & 352k & 18k & 5.5k & 1k \\ \hline
UN & & & & & & & \cellcolor{Gray}{16M} & 3.7k & 3.3k \\ \hline
Wiki Titles & & & & & & & & 118k & 4 \\
\end{tabular}
}
\vspace{-0.5em}
\end{center}
\caption{
Number of training examples with identical English sides split by data sources. E.g.
cell cc-CzEng shows the number of training examples with identical English side by only considering training data coming from either commoncrawl or CzEng for all language pairs (if available). 
}
\label{table:wmt_domain_overlap}
\vspace{-1em}
\end{table*}

Some of the extracted non-English training examples are multi-way parallel by construction. 
The UN corpus is a 6-way parallel corpus, and three of the languages (English, French and Spanish) are in our 6-language mix. A portion of the Europarl corpus is again multi-way aligned. Nevertheless, a good amount of the extracted data is coming from different sources.
Table~\ref{table:wmt_domain_overlap} shows the number of non-English bilingual training examples separated by the two sources they originated from.

\begin{figure}[ht]
\begin{center}
\includegraphics[width=0.5\textwidth]{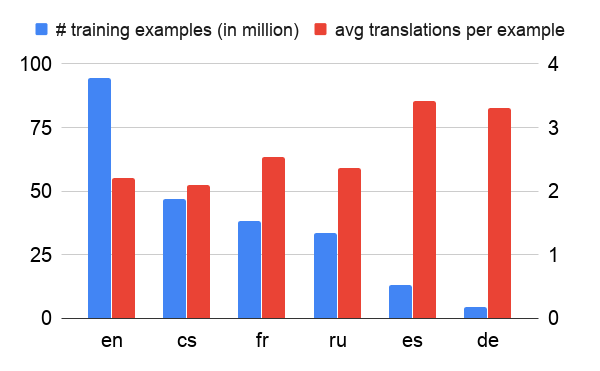}
\caption{Average translations per multi-way aligned example conditioned on the target language.}
\label{fig:wmt_avg_translations}
\end{center}
\vspace{-1em}
\end{figure}

Table~\ref{table:wmt_nway_stats} shows how many translations are available for each sentence in the WMT training data. The majority (123 million) of the multi-way aligned examples do only have translations into two languages. As our original bilingual training data is English-centric, all of the 123 million training examples consist of an English sentence and a translation into one of our five other languages. A total of 13 million multi-way aligned examples are available in at least three languages. Further, Figure~\ref{fig:wmt_avg_translations} shows the average number of translations conditioned by the language. Both Spanish and German have, on average more than three translations. In comparison, the majority of the multi-way aligned examples with Czech or English on the target side are bilingual (having only two translations). Our study resurfaced the inherent multi-way aligned information in the commonly used set of parallel corpora instead of discarding this information.

\begin{table}[ht]
\begin{center}
{\setlength{\tabcolsep}{.3em}
\begin{tabular}{ l||c|c|c|c|c}
\# languages & 2 & 3 & 4 & 5 & 6 \\ \hline \hline
training data & 123M & 6.9M & 5.4M & 0.7M & 10k  \\
\end{tabular}
}
\vspace{-0.5em}
\end{center}
\caption{Data statistics for the extracted multi-way aligned training examples for WMT: 123 million sentences are only available in 2 languages, while 10,000 sentences have translations in all 6 languages.}
\label{table:wmt_nway_stats}
\vspace{-1em}
\end{table}

\section{Complete Multilingual NMT}
\label{sec:extract_nway}

We call MNMT models that are trained for all possible source--target pairs as complete MNMT as all languages are connected via training data (also see Figure~\ref{fig:graph}). Before going into details of how the missing pairs' data gathered, we recap MNMT first.

\paragraph{Multilingual NMT Framework}
MNMT \cite{firat2016zero,johnson2017google} is an extension of bilingual NMT which uses a single model to translate between multiple languages. The model parameters are trained on a joint set of bilingual corpora from different language pairs. Given the data imbalance across the different corpora, it is common to oversample the language pairs with less training data \cite{DBLP:journals/corr/LeeCH16,johnson2017google}. For a given language pair $p$, let $D(p)$ be the size of the available parallel corpus, the sample probability with a temperature $T$ is defined as
\begin{equation}
\label{eqn:sampling}
p_{p} = {(\frac{D(p)}{\sum_{q} D(q)})}^{\frac{1}{T}}
\end{equation}
As a result, $T=1$ corresponds to the actual data distribution, and $T=100$ corresponds to (almost) an equal number of samples for each language pair).
In addition to being able to translate language pairs that the model was trained with, the model can also translate between language pairs never seen explicitly during training which is often referred as zero-shot translation \cite{johnson2017google,ha2016toward}.

\paragraph{Using multi-way aligned data in MNMT}
\label{subsec:multiway_nmt}

Instead of only relying on bilingual corpora, bilingual examples from different language pairs with identical target sentences can be combined into a single multi-way aligned training example. An example is given in Table~\ref{fig:en-pivot}. By comparing the English sides of the Spanish--English and the German--English corpora, we extract a multi-way aligned example that contains translations into all three languages.

\begin{table}[ht]
\centering
\small
\begin{tabular}{c|l}
X-Y &\textcolor{red}{\textit{Bleib sicher}} $\leftrightarrow$ \underline{Stay safe} \\
Z-Y &\textcolor{blue}{Mantente segura} $\leftrightarrow$ \underline{Stay safe}\\
X-Y-Z &\textcolor{red}{\textit{Bleib sicher}} $\leftrightarrow$ \textcolor{blue}{Mantente segura} $\leftrightarrow$ \underline{Stay safe}\\
\end{tabular}
\vspace{-0.5em}
\caption{The two German--English and Spanish--English bilingual training examples can be combined into one multi-way aligned training example that consists of translations into all three languages.}
\label{fig:en-pivot}
\vspace{-1em}
\end{table}

While we can extract direct training data for any source-target pair among the languages considered, the total number of language pairs increases quadratically. The vanilla language pair based sampling strategy in Eq.~(\ref{eqn:sampling}) with adjustable temperature is capable of balancing low-high resource language pairs during training. However, we noticed a critical failure mode, which is further amplified in complete MNMT. 
The language-pair based sampling strategy (regardless of the temperature being used) \textit{over-represents English} in English-centric models. Notice half of the languages have English on the source side, with the other half on the target side. This over-representation yields a schedule of examples for the encoder (resp. for the decoder) to see English examples half of the time throughout the entire training process. As a result, trained models end up favouring English either on the source and/or target. Although the implications on the encoder could be minimal, over-exposing English examples to the decoder curtail the learning signal when the target language is non-English. We hypothesise that this imbalance in the learning signal with respect to the target language is one of the roots of poor translation quality of multilingual models when translating out of English \cite{firat2016multi,johnson2017google,DBLP:journals/corr/abs-1907-05019}. 

To alleviate the \textit{over-representation of English} with the language-pair based sampling strategy, we propose a hierarchical sampling strategy with two levels: i) we choose a target language (based on a temperature-based schedule), ii) uniformly sample a source language.
Formally, for a given target language $l$, let $D(l)$ be the size of the available training examples with target language $l$, the sample probability with a temperature $T$ is defined as
\begin{equation}
\label{eqn:sampling_multiway}
p_{l} = {(\frac{D(l)}{\sum_{q} D(q)})}^{\frac{1}{T}}
\end{equation}
During training, the scheduler samples a batch of training examples based on the target language only, as opposed to source-target language pair specific sampling. After choosing a target language, for each multi-way aligned example, we randomly (uniformly) pick one of the translations as the source sentence.

\section{Experiments}
\label{exper_results}
We use a public transformer implementation 
with the transformer-big model size~\cite{vaswani2017attention} for all multilingual setups. 
All bilingual models use a vocabulary of 32,000 subwords, while all multilingual models use a vocabulary of 64,000 subword units.
All multilingual models are trained for 500,000 updates using an average batch size of around 33,000 sentences ($\sim$1 million tokens). All bilingual models are trained for 400,000 steps as they converged earlier using a batch size of around 8,000 sentences ($\sim$260,000 tokens).
Due to the data imbalance across languages, we use a temperature-based data sampling strategy to over-sample low-resource language pairs in standard MNMT models (Equation~\ref{eqn:sampling}) and low-resource target languages in cMNMT models (Equation~\ref{eqn:sampling_multiway}). We use a temperature of $T$ = 5 in both cases.
All multilingual models add a token at the beginning of the input sentence to specify the required target language.
All BLEU~\cite{papineni2002bleu} scores are calculated with sacreBLEU~ \cite{post2018call}.\footnote{sacreBLEU signatures: BLEU+case.mixed+lang.SRC-TGT+numrefs.1+smooth.exp+SET+tok.intl+version.1.2.20}

\subsection{Baselines on WMT}
\label{subsec:wmt_baselines}
We train several baselines: (i) bilingual models, (ii) multilingual models based on English-centric data, and (iii) bridging non-English language pairs.

\paragraph{Bilingual Baselines}
We train two bilingual baselines (using either transformer-base or transformer-big) for each language pair. In addition to training baselines on the original English-centric WMT data, we also train models for non-English language pairs on the extracted direct data (see Table~\ref{table:wmt_training_data_stats}). We experimented with several dropout rates for both setups and found that dropout=0.1 works best for transformer-base while dropout=0.3 works best for transformer-big. 
As can be seen from Table~\ref{table:wmt_bilingual_baseline_small} and Table~\ref{table:wmt_bilingual_baseline_big}, the experiments suggest that the translation quality of the non-English language pairs is far behind the ones for English-centric language pairs. As an example, the translation quality between German and Russian reaches 6-7 BLEU only.

\begin{table}[!ht]
\begin{center}
{\setlength{\tabcolsep}{.45em}
\begin{tabular}{ cl||c|c|a|c|c|c}
\multicolumn{2}{c}{~} & \multicolumn{6}{c}{target} \\
\multirow{8}{*}{\rotatebox[origin=c]{90}{source}} & & cs & de & \cellcolor{white}{en} & es & fr & ru\\ \cline{2-8} \cline{2-8}
 & cs & & 16.6 & 30.4 & 20.7 & 22.6 & 13.9 \\ \cline{2-8}
 & de & 15.2 & & 29.5 & 27.0 & 28.7 & 6.9 \\ \cline{2-8}
& en & \cellcolor{Gray}{25.2} & \cellcolor{Gray}{25.7} & & \cellcolor{Gray}{33.6} & \cellcolor{Gray}{34.8} & \cellcolor{Gray}{23.6} \\ \cline{2-8}
 & es & 15.6 & 22.7 & 33.9 & & 34.2 & 18.7 \\ \cline{2-8}
 & fr& 15.4 & 22.1 & 33.0 & 31.8 & & 17.9 \\ \cline{2-8}
 & ru & 12.5 & 6.1 & 28.4 & 22.6 & 24.3 & \\
\end{tabular}
}
\end{center}
\vspace{-0.5em}
\caption{BLEU scores on newstest2013 of bilingual models trained with the transformer-base architecture.}
\label{table:wmt_bilingual_baseline_small}
\vspace{-1em}
\end{table}

\begin{table}[!ht]
\begin{center}
{\setlength{\tabcolsep}{.45em}
\begin{tabular}{ cl||c|c|a|c|c|c}
\multicolumn{2}{c}{~} & \multicolumn{6}{c}{target} \\
\multirow{8}{*}{\rotatebox[origin=c]{90}{source}} & & cs & de & \cellcolor{white}{en} & es & fr & ru\\ \cline{2-8} \cline{2-8}
 & cs & & 14.6 & 31.9 & 19.0 & 20.0 & 14.1 \\ \cline{2-8}
 & de & 14.1 & & 31.3 & 26.4 & 28.8 & 4.7 \\ \cline{2-8}
& en & \cellcolor{Gray}{26.5} & \cellcolor{Gray}{27.0} & & \cellcolor{Gray}{34.2} & \cellcolor{Gray}{35.9} & \cellcolor{Gray}{25.0} \\ \cline{2-8}
 & es & 14.4 & 22.8 & 34.5 & & 34.8 & 19.9 \\ \cline{2-8}
 & fr & 13.0 & 20.7 & 34.2 & 32.5 & & 18.6 \\ \cline{2-8}
 & ru & 12.8 & 4.0 & 30.8 & 23.1 & 24.8 & \\
\end{tabular}
}
\end{center}
\vspace{-0.5em}
\caption{BLEU scores on newstest2013 of bilingual models trained with the transformer-big architecture.}
\label{table:wmt_bilingual_baseline_big}
\vspace{-1.2em}
\end{table}

\paragraph{Multilingual Baselines}

We train a multilingual NMT model on the original WMT English-centric training data.
BLEU scores are summarized in Table~\ref{table:wmt_en_central_bleu}. All language pairs with English as the source or target language perform comparably well from at least 24.5 BLEU (English$\to$Russian) up to 34.9 BLEU (English$\to$French). 
The BLEU scores of non-English language pairs are consistently lower (which can be explained as a lack of supervision during training) and can be as low as 4.1 BLEU for Spanish$\to$Czech or as high as 24.4 BLEU for French$\to$Spanish.

\vspace{-0.7em}
\begin{table}[!ht]
\begin{center}
{\setlength{\tabcolsep}{.45em}
\begin{tabular}{ cl||c|c|a|c|c|c}
\multicolumn{2}{c}{~} & \multicolumn{6}{c}{target} \\
\multirow{8}{*}{\rotatebox[origin=c]{90}{source}} & & cs & de & \cellcolor{white}{en} & es & fr & ru\\ \cline{2-8} \cline{2-8}
& cs & & 19.8 & 31.2 & 21.6 & 20.2 & 8.5 \\ \cline{2-8}
& de & 6.8 & & 31.8 & 17.8 & 21.2 & 4.5 \\ \cline{2-8}
& en & \cellcolor{Gray}{25.5} & \cellcolor{Gray}{26.7} & & \cellcolor{Gray}{34.0} & \cellcolor{Gray}{34.9} & \cellcolor{Gray}{24.5} \\ \cline{2-8}
& es & 4.1 & 8.8 & 34.7 & & 19.6 & 9.5 \\ \cline{2-8}
& fr & 4.2 & 11.2 & 33.8 & 24.4 & & 6.5 \\ \cline{2-8}
& ru & 4.8 & 10.4 & 29.5 & 19.9 & 9.6 & \\ 
\end{tabular}
}
\end{center}
\vspace{-0.5em}
\caption{BLEU scores on newstest2013 of a MNMT model trained on English-centric training data. All non-English language pairs are unseen during training and BLEU scores measure zero-shot performance.}
\label{table:wmt_en_central_bleu}
\vspace{-1em}
\end{table}

\paragraph{Bridging (Pivoting) Baselines}
The quality of MNMT is still behind the one from bilingual baselines for most of the language pairs (comparing Table~\ref{table:wmt_bilingual_baseline_big} and Table~\ref{table:wmt_en_central_bleu}). Nevertheless, having a single NMT model for each language pair is impractical, especially when increasing the number of language pairs. An alternative approach is called bridging \cite{cohn-lapata-2007-machine,wu2007pivot,utiyama2007comparison}. For the bridging approach, we compromise and train only English-centric models. To enable the translation between non-English language pairs, the source sentence cascades through the source$\to$English and English$\to$target systems to generate the target sentence. This simple process has several limitations: (i) translation errors accumulate in the pipeline, (ii) decoding time gets doubled since inference has to be run twice, (iii) bridging through a morphologically low language (i.e. English), important information could be lost (i.e. gender). 
The BLEU scores (Table~\ref{table:wmt_bridging}) for all non-English pairs are higher compared to all previous baselines. We can reach acceptable translation quality even for German$\to$Russian, where our direct training data is scarce. We use the bridging baseline to compare our cMNMT models in the rest of the paper.

\begin{table}[!ht]
\begin{center}
{\setlength{\tabcolsep}{.45em}
\begin{tabular}{ cl||c|c|a|c|c|c}
\multicolumn{2}{c}{~} & \multicolumn{6}{c}{target} \\
\multirow{8}{*}{\rotatebox[origin=c]{90}{source}} & & cs & de & \cellcolor{white}{en} & es & fr & ru\\ \cline{2-8} \cline{2-8}
& cs & & 22.4 & 31.9 & 27.0 & 28.8 & 21.9\\ \cline{2-8}
& de & 21.5 & & 31.3 & 26.9 & 29.0 & 20.3 \\ \cline{2-8}
& en & \cellcolor{Gray}{26.5} & \cellcolor{Gray}{27.0} & & \cellcolor{Gray}{34.2} & \cellcolor{Gray}{35.9} & \cellcolor{Gray}{24.9} \\ \cline{2-8}
& es & 22.6 & 22.8 & 34.5 &  & 32.6 & 22.5 \\ \cline{2-8}
& fr & 21.4 & 22.2 & 34.2 & 29.1 & & 21.6 \\ \cline{2-8}
& ru & 21.3 & 20.6 & 30.8 & 27.2 & 28.5 & \\ 
\end{tabular}
}
\end{center}
\vspace{-0.5em}
\caption{BLEU scores on newstest2013 for our WMT setup. Translations for non-English language pairs are generated via bridging over English.}
\label{table:wmt_bridging}
\vspace{-1.2em}
\end{table}

\subsection{Complete MNMT Models on WMT}

Without adding new training data and taking into account the multi-way property of the data, we train a complete multilingual NMT system (cMNMT, see Section~\ref{subsec:multiway_nmt}).
We compare the performance of cMNMT with the best baseline model that is based on bridging (Table~\ref{table:wmt_bridging}) and report BLEU and delta BLEU numbers in Table~\ref{table:wmt_multiway_bleu}. The BLEU scores for the non-English language pairs go up from at least 1.4 BLEU for Russian$\to$Spanish up to 5.0 BLEU for Czech$\to$Russian.
We changed the sampling strategy for our cMNMT models to be conditioned on the target language only (Section~\ref{subsec:multiway_nmt}). As a result, English has been seen less often as the target language when compared to a standard MNMT setup. Interestingly, this seems to affect only the performance of Russian$\to$English, which shows a decrease of 1 BLEU point. The other language pairs with English as the target language are keeping their translation quality.

\begin{table}[ht]
\begin{center}
{\setlength{\tabcolsep}{.45em}
\begin{tabular}{ cl||c|c|a|c|c|c}
\multicolumn{2}{c}{~} & \multicolumn{6}{c}{target} \\
\multirow{12}{*}{\rotatebox[origin=c]{90}{source}} & & cs & de & \cellcolor{white}{en} & es & fr & ru\\ \cline{2-8} \cline{2-8}
 & cs & & 25.8 & 32.0 & 30.1 & 31.4 & 26.9 \\ [-0.35em]
  &  & & \scell{+3.4} & \sgcell{+0.1} & \scell{+3.1} & \scell{+2.6} & \srcell{+5.0} \\ \cline{2-8}
 & de & 23.9 & & 31.2 & 29.9 & 31.8 & 23.4 \\ [-0.35em]
 &  & \scell{+2.4} & & \sgcell{-0.1} & \scell{+3.0} & \scell{+2.8} & \srcell{+3.1} \\ \cline{2-8}
 & en & \cellcolor{Gray}{26.9} & \cellcolor{Gray}{27.1} & & \cellcolor{Gray}{35.0} & \cellcolor{Gray}{35.5} & \cellcolor{Gray}{26.4} \\ [-0.35em]
  & & \sgcell{+0.4} & \sgcell{+0.1} & & \sgcell{+0.8} & \sgcell{-0.4} & \sgrcell{+1.5} \\ \cline{2-8}
 & es & 24.9 & 25.7 & 34.9 & & 36.0 & 24.9 \\ [-0.35em] 
 & & \scell{+2.3} & \scell{+2.9} & \sgcell{+0.4} & & \scell{+3.4} & \srcell{+2.4} \\ \cline{2-8}
 & fr & 23.7 & 25.2 & 34.2 & 33.3 & & 23.5 \\ [-0.35em]
  &  & \scell{+2.3} & \scell{+3.0} & \sgcell{+0.0} & \scell{+4.2} & & \srcell{+1.9} \\ \cline{2-8}
 & ru & 24.3 & 22.7 & 29.8 & 28.6 & 30.1 & \\ [-0.35em]
  & & \scell{+3.0} & \scell{+2.1} & \sgcell{-1.0} & \scell{+1.4} & \srcell{+1.6} & \\ 
\end{tabular}
}
\end{center}
\vspace{-0.5em}
\caption{BLEU on newstest2013 for our novel cMNMT model. The small numbers are the difference ($\Delta$BLEU) with respect to the bridging approach (Table~\ref{table:wmt_bridging}).}
\label{table:wmt_multiway_bleu}
\vspace{-1em}
\end{table}

When comparing our cMNMT model to the English-centric baseline (Table~\ref{table:wmt_en_central_bleu}), we see an average BLEU increase of 14.6 BLEU for all non-English language pairs.
It is worth noticing that every language pair has now at least 22 absolute BLEU points. Interestingly, the absolute BLEU scores in each row (translations into the same language) are much closer, suggesting a more universal input representation.

\section{Analysis and Discussion}
\label{subsec:add_exp}

To further understand the impact of multi-way aligned examples on NMT, we run a couple of additional experiments. 

\begin{figure}[ht]
\begin{center}
\includegraphics[width=0.5\textwidth]{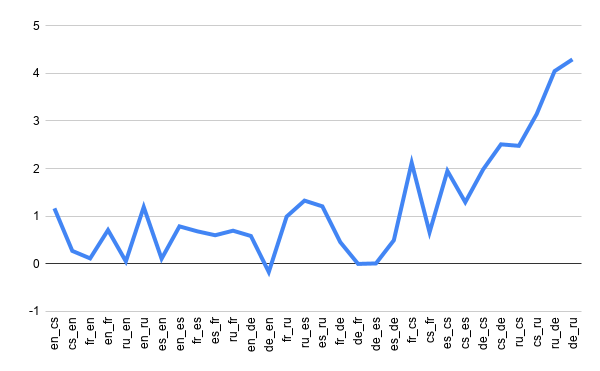}
\vspace{-2em}
\caption{$\Delta$BLEU scores for a target language based versus a language-pair based temperature schedule. }
\label{fig:temp_comp}
\end{center}
\vspace{-1em}
\end{figure}

\paragraph{Training Data Sampling Strategy}
In Section~\ref{subsec:multiway_nmt}, we did introduce our new training data sampling strategy that is based on the target language only. This change was mainly driven by the fact that having a language-pair conditioned schedule is not scalable when building a system of 12,432 language pairs. Instead of finding a good sampling weight for each of the 12,432 language pairs, we only need to find a suitable mix for the 112 target languages. Further, we have more control over how often each target language will be seen during training. To see the impact of this change, we train an MNMT system on the joint set of the 30 different bilingual corpora with a standard language-pair based temperature scheduling scheme and compare it to a cMNMT model. We used temperature 5 in both setups. $\Delta$BLEU numbers for each language-pair can be seen in Figure~\ref{fig:temp_comp}. The language-conditioned temperature scheduling increases BLEU scores for 29 out of 30 language-pairs with larger gains for the low-resource language-pairs. This experiment suggests that a target language based temperature scheduling is not only simpler but also performs better on average.

\paragraph{Separate Multi-way Aligned Examples}
We test the transfer learning capability of cMNMT by
training a cMNMT model only on the 13 million multi-way aligned examples that have translations in at least three languages (see Table~\ref{table:wmt_nway_stats}). 
In other words, we remove all training examples that are only available in English and one additional language. If no transfer learning is happening, the English-centric scores will decrease while the BLEU numbers of the non-English language pairs are not affected. 
Experimental results can be seen in Table~\ref{table:wmt_nwayonly_bleudiff}. Interestingly, we find that the performance of all language pairs is similarly affected. This indicates that transfer learning is happening between the language pairs and that non-English language pairs benefit from having more English-centric data.

\begin{table}[ht]
\begin{center}
{\setlength{\tabcolsep}{.45em}
\begin{tabular}{ cl||c|c|a|c|c|c}
\multicolumn{2}{c}{~} & \multicolumn{6}{c}{target} \\
\multirow{12}{*}{\rotatebox[origin=c]{90}{source}} & & cs & de & \cellcolor{white}{en} & es & fr & ru\\ \cline{2-8} \cline{2-8}
 & cs & & 24.1 & 30.3 & 28.4 & 29.4 & 23.6 \\ [-0.35em]
  &  & & \scell{-1.7} & \sgcell{-1.7} & \scell{-1.7} & \scell{-2.0} & \srcell{-3.3} \\ \cline{2-8}
 & de & 19.1 & & 30.6 & 29.1 & 30.5 & 21.7 \\ [-0.35em]
&  & \scell{-4.8} & & \sgcell{-0.6} & \scell{-0.8} & \scell{-1.3} & \srcell{-1.7} \\ \cline{2-8}
 & en & \cellcolor{Gray}{21.8} & \cellcolor{Gray}{26.2} & & \cellcolor{Gray}{33.4} & \cellcolor{Gray}{34.6} & \cellcolor{Gray}{23.3} \\ [-0.35em]
  & & \sgcell{-5.1} & \sgcell{-0.9} & & \sgcell{-1.6} & \sgcell{-0.9} & \sgrcell{-3.1} \\ \cline{2-8}
 & es & 19.7 & 24.9 & 34.3 & & 35.1 & 22.7 \\ [-0.35em]
& & \scell{-5.2} & \scell{-0.8} & \sgcell{-0.6} & & \scell{-0.9} & \srcell{-2.2} \\ \cline{2-8}
 & fr & 18.6 & 24.0 & 33.0 & 32.3 & & 21.3 \\ [-0.35em]
 &  & \scell{-5.1} & \scell{-1.2} & \sgcell{-1.2} & \scell{-1.0} & & \srcell{-2.2} \\ \cline{2-8}
 & ru & 19.7 & 21.4 & 27.7 & 27.1 & 27.9 & \\ [-0.35em]
 & & \scell{-4.6} & \scell{-1.3} & \sgcell{-1.9} & \scell{-1.5} & \srcell{-2.2} & \\ 
\end{tabular}
}
\vspace{-0.5em}
\end{center}
\caption{BLEU on newstest2013 for a model trained on 13 million multi-way aligned ($n$$>$$2$) data. Small numbers are the difference ($\Delta$BLEU) between cMNMT trained on all multi-way examples (136M, Table~\ref{table:wmt_multiway_bleu}).}
\label{table:wmt_nwayonly_bleudiff}
\vspace{-0.5em}
\end{table}

To further study this effect, we reverse that experiment and remove all examples that have translations into more than two languages.
This experiment investigates if the non-English language pairs in a standard MNMT model can benefit from having training examples with identical English sides.
Experimental results can be found in Table~\ref{table:wmt_2wayonly_bleudiff}. The BLEU scores for English-centric language pairs drop by 0.9 points on average while the performance of non-English language pairs decreases by 1.6 BLEU on average.

\begin{table}[ht]
\begin{center}
{\setlength{\tabcolsep}{.45em}
\begin{tabular}{ cl||c|c|a|c|c|c}
\multicolumn{2}{c}{~} & \multicolumn{6}{c}{target} \\
\multirow{12}{*}{\rotatebox[origin=c]{90}{source}} & & cs & de & \cellcolor{white}{en} & es & fr & ru\\ \cline{2-8} \cline{2-8}
 & cs & & 17.8 & 31.5 & 14.5 & 20.3 & 5.2 \\ [-0.35em]
 &  & & \scell{-2.0} & \sgcell{+0.3} & \scell{-7.1} & \scell{+0.1} & \srcell{-3.3} \\ \cline{2-8}
 & de & 7.8 & & 29.6 & 17.2 & 22.9 & 1.8 \\ [-0.35em]
 &  & \scell{+1.0} & & \sgcell{-2.2} & \scell{-0.6} & \scell{+1.7} & \srcell{-2.7} \\ \cline{2-8}
 & en & \cellcolor{Gray}{25.6} & \cellcolor{Gray}{23.9} & & \cellcolor{Gray}{33.1} & \cellcolor{Gray}{33.5} & \cellcolor{Gray}{24.6} \\ [-0.35em]
  & & \sgcell{+0.1} & \sgcell{-2.8} & & \sgcell{-0.9} & \sgcell{-1.4} & \sgrcell{+0.1} \\ \cline{2-8}
 & es & 7.2 & 3.9 & 32.7 & & 24.3 & 7.6 \\ [-0.35em]
 & & \scell{+3.0} & \scell{-5.8} & \sgcell{-1.1} & & \scell{+4.7} & \srcell{-1.9} \\ \cline{2-8}
 & fr & 6.7 & 13.0 & 33.1 & 19.3 & & 7.2 \\ [-0.35em]
  &  & \scell{-2.5} & \scell{-1.8} & \sgcell{-0.7} & \scell{-5.1} & & \srcell{+0.7} \\ \cline{2-8}
 & ru & 5.5 & 10.4 & 29.3 & 8.5 & 13.7 & \\ [-0.35em]
  & & \scell{+0.7} & \scell{+0.0} & \sgcell{-0.2} & \scell{-14.4} & \srcell{+4.1} & \\ 
\end{tabular}
}
\vspace{-0.5em}
\end{center}
\caption{BLEU on newstest2013 for a model trained on 2-way data only. Small numbers are the difference ($\Delta$BLEU) between the vanilla MNMT model (Table~\ref{table:wmt_en_central_bleu}).}
\label{table:wmt_2wayonly_bleudiff}
\vspace{-1em}
\end{table}

\begin{figure}[ht]
\begin{center}
\includegraphics[width=0.5\textwidth]{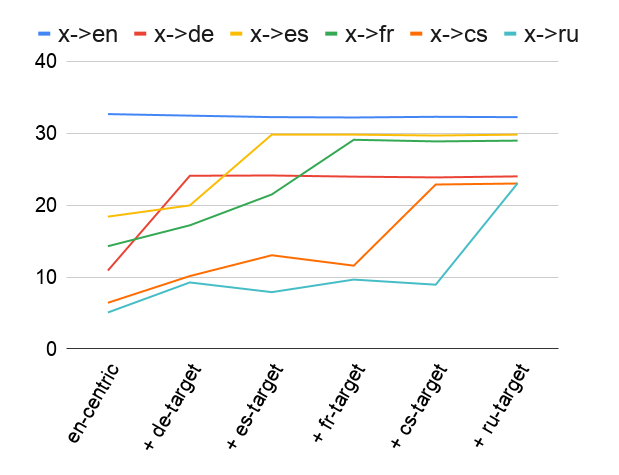}
\vspace{-2em}
\caption{BLEU scores of models using only parts of the multi-way training data. }
\label{fig:leave_some_out}
\end{center}
\vspace{-1em}
\end{figure}

\paragraph{Leave N-Out}
\label{subsec:leave1out}
We further investigate the transfer learning capability of our approach by training several cMNMT models on different amounts of training data. We start with a cMNMT model trained on English-centric bilingual training data only. This setup ensures that all languages have been seen on both the source and target side during training. We further group the remaining multi-way aligned training examples by target language and add one after another to the training data. Important to mention: We retrained all configurations from scratch. Experimental results are summarized in Figure~\ref{fig:leave_some_out}. We report average BLEU scores grouped by the target language. We can see that adding training data $x \to y$ for a target language $y$, gives a significant boost in translation quality for that target language. These results demonstrate that even though we can translate between language pairs without seeing a single example during training, adding supervision during training significantly increases BLEU scores.

\paragraph{Adding a New Language}
\label{subsec:fine_tuning}
We further investigate how a cMNMT model behaves when fine-tuned \cite{freitag2016fast} to a new language.
We chose Italian as the new language as the test set newstest2009 is multi-way in Czech, English, French, German, Italian and Spanish and thus we can report BLEU scores between all language pairs. 
We run two experiments with two different sets of fine-tuning data. First, we fine-tuned the cMNMT model (Table~\ref{table:wmt_multiway_bleu}) on English$\leftrightarrow$Italian news-commentary (45,000 examples). 
Second, we converted the same data into multi-way aligned examples by augmenting the bilingual examples with translations into other languages when found in our original training data. 
Experimental results for fine-tuning our model for one epoch on either of the two datasets can be found in
 Table~\ref{table:wmt_finetune_encentric}. Both fine-tuning experiments show the same BLEU improvements for Italian$\leftrightarrow$English. Nevertheless, when only fine-tuning on English$\leftrightarrow$Italian data, we sacrifice translation quality for most of the language pairs which can be seen in the $x \to y$ column. Further, fine-tuning on multi-way aligned examples does improve the average BLEU scores by 4.3 BLEU for translations into Italian ($x \to$it). Overall, these experiments suggest that fine-tuning with multi-way aligned data is superior.  

\begin{table}[ht]
\begin{center}
{\setlength{\tabcolsep}{.3em}
\begin{tabular}{ l||c|c|c|c|c}
model & it$\to$en & it$\to x$ & en$\to$it & $x \to $it & $x \to y$ \\ \hline
cMNMT  & 13.5 & 9.7 & 2.3 & 2.6 & 22.0 \\
\ +ft en$\leftrightarrow$it & 21.5 & 14.2 & 13.6 & 11.8 & 17.8 \\
\ +ft mway & 21.2 & 18.5 & 13.5 & 11.9 & 23.0 \\
\end{tabular}
}
\end{center}
\vspace{-0.5em}
\caption{BLEU scores for newstest2009 for fine-tuning (ft) our cMNMT model on either English$\leftrightarrow$Italian (it$\leftrightarrow$en) news-commentary or on the same sentences but augmented with translations into other languages (mway), if available. Column $x \to y$ shows average BLEU scores for all language pairs.}
\label{table:wmt_finetune_encentric}
\vspace{-1em}
\end{table}

\begin{figure}[ht]
\begin{center}
\includegraphics[width=0.5\textwidth]{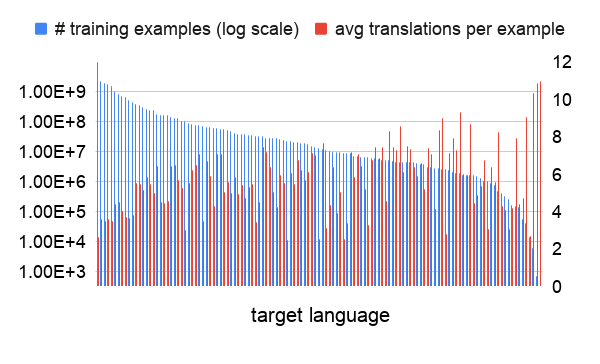}
\vspace{-3em}
\caption{Average translations per multi-way example conditioned on the target language.}
\label{fig:m4_avg_translations}
\end{center}
\vspace{-1em}
\end{figure}

\paragraph{Scaling cMNMT: 12,432 Language Pairs}
We run additional experiments on a 112 language in-house dataset~\cite{DBLP:journals/corr/abs-1907-05019} 
to see if our approach scales to 12,432 language pairs. Our in-house dataset does not only contain more languages than the WMT setup, but also has a much wider range of available training resources. While for the high resource languages, we have access to billions of training examples, most of the low resource languages have less than 1 million training examples. We refer the reader to the description in \newcite{DBLP:journals/corr/abs-1907-05019} for more details regarding the dataset. 
Figure~\ref{fig:m4_avg_translations} shows the training data sizes and the average translations per multi-way example.

Although a deeper and wider architecture does improve the quality of multilingual models for this dataset, we use the same experimental setup as used in our WMT experiments (see Section~\ref{exper_results}) to run an MNMT and cMNMT model on our in-house data.
Experimental results can be seen in Table~\ref{table:m4_results_bleudiff}. cMNMT outperforms MNMT for non-English languages by 10.1 BLEU points on average while keeping the translation quality for language pairs that include English as source or target. These results demonstrate that our proposed approach does scale far behind the six language WMT setup.

\begin{table}[ht]
\begin{center}
{\setlength{\tabcolsep}{.38em}
\begin{tabular}{l|c|c|c}
& en$\to x$ & $x \to$en & $x \to y$ \\ \hline \hline
all & +0.34 & -0.05 & +10.1 \\ \hline
low resource & +0.23 & -0.15 & +8.82 \\ \hline
mid resource & +0.35 & -0.05 & +11.02 \\ \hline
high resource & +0.45 & +0.04 & +9.73 \\
\end{tabular}
}
\end{center}
\vspace{-0.5em}
\caption{Average BLEU difference ($\Delta$BLEU) between a cMNMT and a vanilla MNMT model for our in-house 112 language setup. Positive numbers present improvement of cMNMT over MNMT.}
\label{table:m4_results_bleudiff}
\vspace{-1.5em}
\end{table}

\section{Related Work}
\paragraph{Direct models}
To translate between languages with little training data,
three general approaches emerged, i. bridging through a third language (pivot-based MT) \cite{DBLP:journals/corr/ChengLYSX16,currey2019zero}, ii. generating pseudo-parallel data between direct language pairs and training the direct pairs with that (zero-resource MT) \cite{firat2016zero,chen2017teacher} and, iii. zero-shot methods where the model is asked to translate a direct pair only at test time \cite{johnson2017google, ha2016toward,DBLP:journals/corr/abs-1903-07091}. 

Although pivot-based approaches perform sufficiently good when cascaded with strong bilingual models \cite{DBLP:journals/corr/abs-1906-01181}, their practicality is limited due to compounding errors from pipelining and doubled inference cost.
The zero-resource approaches, combined with iterative-back translation \cite{hoang-etal-2018-iterative} are quite powerful but their inefficiency is worth noting. For $N$ languages, one needs to devise a training routine that could sample $N^2-N$ pairs, generate pseudo-parallel data. The added time to generate pseudo-parallel data for every pair grows quadratically, making it challenging for systems considering a large number of languages. Recently, by devising a practical sub-sampling approach, \citep{zhang2020improving} demonstrated zero-resource techniques could be scaled to massively multilingual setup. We find the study by \cite{zhang2020improving} closest to our work, having the goal of any-to-any multilingual translation. But compared to sampling language pairs with no parallel data and generating pseudo-parallel data on-the-fly, our approach makes use of existing multi-way alignment information before training. Lastly, zero-shot approaches attempt to measure the generalization performance of the MNMT models, but to date, the zero-shot quality still trails behind the pivot and zero-resource methods \cite{al2019consistency}. Our proposed cMNMT, naturally fills the gap between these three approaches, the multi-way data can be extracted offline, and efficiently be mixed with the original data using a hierarchical data sampler. It does not require extra steps to generate pseudo-parallel data, and (as expected) it handily outperforms zero-shot approaches. 

\paragraph{N-way data} In this paper, we only made use of multi-way aligned data to sample bilingual pairs out of it. But there exist several approaches that make use of the multi-view structure in the data, such as \newcite{dabre2019exploiting}, who explored the use of small multi-parallel corpora a for one-to-many NMT. Another approach is multi-source NMT \cite{Zoph_2016}. Although multi-source NMT is a promising direction, it has practical problems such as lacking multiple sources at inference time \cite{nishimura-etal-2018-multi}. We believe research in this direction will be the key to improve mid/high-resource NMT and address several robustness issues to the input noise. 
\newcite{aulamo2020opustools} recently released MultiParaCrawl where the authors extracted direct data for non-English language pairs from the English-centric Paracrawl corpus.

\paragraph{Sampling scheduling} 
Several approaches proposed to address data sampling for multi-task models, some relying on temperature-based heuristics \cite{DBLP:journals/corr/LeeCH16,DBLP:journals/corr/abs-1810-04805,DBLP:journals/corr/abs-1907-05019}, others relying on adaptive schedules that incorporate the model gains, baselines or quality expectations into the data schedulers \cite{kiperwasser2018scheduled,jean2019adaptive,wang2020balancing}. We believe data sampling is a critical research area for not only MNMT but also multi-task learning in general. We reveal a critical failure mode of the commonly used temperature sampling strategy, and how it causes the poor translation quality while translating out of English. 

\section{Conclusion}
In this work, we introduced complete Multilingual Neural Machine Translation (cMNMT) that exploits the multi-way alignment information in the underlying training data to improve translation quality for language pairs where training data is scared or not available. Standard MNMT models are trained on a joint set of different training corpora for a variety of language pairs. cMNMT combines the different corpora and constructs multi-way aligned training examples that consist of translations of the same sentence into multiple languages.
In combination with a novel temperature-based sampling approach that is conditioned on the target language only, we show that cMNMT is superior to the standard MNMT model and the even better-performing bridging approach. 
Experimental results on a public WMT 30 language pairs dataset and an in-house 12,432 language pairs dataset demonstrated an average BLEU increase of more than 10 BLEU points for non-English language pairs. This approach leads to a single NMT model that can serve 12,432k language pairs with reasonable quality which also surpasses the translation quality of the bridging approach, which is nowadays used in most modern MT services.

\bibliography{emnlp2020}
\bibliographystyle{acl_natbib}

\appendix

\end{document}